\documentclass[runningheads]{llncs}

\usepackage{graphicx}
\usepackage{booktabs}
\usepackage{comment}

\usepackage{amsmath}
\usepackage{amssymb}
\usepackage{hyperref}
\usepackage{multirow}
\usepackage{caption}
\usepackage{tikz}
\usetikzlibrary{arrows.meta,positioning,shapes.geometric}
\usepackage{float}

\begin{document}

\title{Medical Knowledge Simplification for Patients in the Era of LLMs: A Case Study on Diabetes}
\author{
Pallika Kafle\inst{1}
\and Yipeng Zhou\inst{1}
\and Guanfeng Liu\inst{1}
\and Quan Z. Sheng\inst{2}
\and Sheng-Hsin Hsu\inst{3}
}

\author{
Pallika Kafle\inst{1}
\and Yipeng Zhou\inst{1}\orcidID{0000-0003-1533-0865}
\and Guanfeng Liu\inst{1}\orcidID{0000-0001-8980-4950}
\and Quan Z. Sheng\inst{2}
\and Cheng-Hsin Hsu\inst{3}\orcidID{0000-0001-7302-6488}
}
\author{
Pallika Kafle\inst{1}
\and
Yipeng Zhou\inst{1}
\orcidID{0000-0003-1533-0865}
\and
Guanfeng Liu\inst{1}
\orcidID{0000-0001-8980-4950}
\and
Quan Z. Sheng\inst{2}
\orcidID{0000-0002-3326-4147}
\and
Cheng-Hsin Hsu\inst{3}
\orcidID{0000-0002-8116-2591}
}

\authorrunning{P. Kafle et al.}

\institute{
Macquarie University, Sydney, Australia
\and
Beijing Normal--Hong Kong Baptist University, Zhuhai, China
\and
National Tsing Hua University, Hsinchu, Taiwan
\email{yipeng.zhou@mq.edu.au}
}


\maketitle

\begin{abstract}

Complex medical information is often difficult for patients to understand, making effective medical knowledge simplification essential for improving patient comprehension, informed decision-making, and health outcomes. Recent advances in large language models (LLMs) provide a promising approach for simplifying complex medical information into patient-friendly language; however, their effectiveness in real-world patient education remains insufficiently explored through human evaluation.
To investigate their practical effectiveness, this paper presents a case study on diabetes knowledge simplification through the implementation and evaluation of \emph{MediClear}, an LLM-based medical knowledge simplification system enhanced with Retrieval-Augmented Generation (RAG). Public diabetes-related articles from Diabetes Australia, WHO, American Diabetes Assiciation (ADA), NIDDK, and AIHW are indexed in the RAG knowledge base to retrieve clinically grounded information, which is then simplified by the LLM into accessible patient explanations. We evaluate the generated responses using standard readability metrics, including the Flesch--Kincaid Grade Level (FKGL), and conduct a human study involving 10 participants. Results show that \emph{MediClear} consistently reduces the reading level of generated responses to the recommended patient literacy range while achieving high user satisfaction and willingness for future use. This case study demonstrates the potential of  LLMs to improve the accessibility of medical knowledge for patient education.

\end{abstract}

\section{Introduction}
\label{sec:intro}

Health literacy is the ability to access, understand, and apply health information to make informed healthcare decisions, encompassing the reading, numeracy, listening, and communication skills needed to navigate the healthcare system and follow clinical instructions~\cite{helitzer2012talkdoc}. However, inadequate health literacy remains widespread across diverse populations~\cite{parker2005outreach,shah2022attitudes,hjorth2025cirrhosis}. Poor health literacy is a stronger predictor of adverse health outcomes than age, income, employment, or education level, and is associated with increased hospitalizations, emergency department revisits, and healthcare costs exceeding \$100 billion annually~\cite{shahid2022impact}.

The widespread use of the internet and social media for health information has not closed this gap and may even have exacerbated it. Many people now rely on online platforms as their primary source of health information~\cite{ghalavand2024socialmedia}, where content is often written beyond the recommended reading level and varies substantially in quality and reliability~\cite{parker2005outreach}. Consequently, individuals with limited health literacy are more vulnerable to misunderstanding or misinformation, highlighting the need for tools that can provide accurate, accessible, and easy-to-understand health information on demand.

Large language models (LLMs) have emerged as a promising tool for simplifying complex medical information into patient-friendly language, and hence improve health literacy. However, directly applying LLMs for enhancing health literacy face two challenges. First, LLMs are prone to hallucinations, which may produce inaccurate or unsafe medical advice~\cite{ahmad2025beyond,moell2025journaling}. RAG addresses this limitation by grounding responses in trusted medical resources~\cite{klesel2025rag,neha2025ragreview}. Second, while prior studies have reported improved readability of LLM-generated text, evaluations have largely relied on automated readability metrics rather than human assessment of patient comprehension~\cite{aydin2024scoping}. As a result, the effectiveness of RAG-enhanced LLMs in producing medical explanations that are both accurate and understandable remains largely unexplored.

This paper investigates the potential of LLMs to practically enhance health literacy through the design and evaluation of \emph{MediClear}, a RAG-enhanced LLM framework for simplifying complex medical information. Based on \emph{MediClear}, we also conduct a case study on diabetes involving human participants to assess its effectiveness.

\emph{MediClear} integrates a RAG module with state-of-the-art LLMs to improve both factual accuracy and readability. Medical knowledge collected from publicly available healthcare resources is indexed into the RAG database~\cite{klesel2025rag,wang2025hybridsearch}, ensuring that generated explanations are grounded in reliable evidence. In addition, we construct a manually curated plain-language medical lexicon that maps complex clinical terminology to patient-friendly expressions. For each user query, relevant medical knowledge is first retrieved from the RAG database and then combined with lexical simplification prompts to guide the LLM in generating accessible explanations. An iterative refinement process evaluates the generated text using the Flesch--Kincaid Grade Level (FKGL); if the target reading level is not achieved, the explanation is regenerated until the readability requirement is satisfied.

To evaluate \emph{MediClear}, we conduct both standard evaluation, using metrics FKGL and the Simple Measure of Gobbledygook (SMOG) index~\cite{readable2025smog}, and a human study with 10 participants, comparing the original medical text with the simplified output. The results show that \emph{MediClear} reduces the reading level by 39.8--57.5\% in terms of mean FKGL score compared with an ungrounded LLM baseline while maintaining high perceived trustworthiness. Furthermore, 91.8\% of collected responses (45 of 49) indicated a willingness to use the system in future, demonstrating its potential to improve health literacy through accurate and accessible medical explanations.

In summary, our work has made the following contributions:
\begin{itemize}
    \item We develop \emph{MediClear}, a RAG-enhanced LLM framework that integrates medical knowledge grounding and controlled language simplification for improving the accessibility of patient education materials.
    
    \item We propose a closed-loop simplification strategy that combines a patient-oriented medical lexicon with readability-aware refinement to automatically generate explanations aligned with target literacy levels.
    
    \item We conduct a human-centered evaluation on diabetes-related medical information, demonstrating that LLM-based simplification can substantially improve readability while preserving perceived accuracy and trustworthiness.
\end{itemize}

The remainder of the paper is organized as follows. Section~\ref{sec:related} reviews related work on health literacy, LLM-based simplification, and RAG. Section~\ref{sec:design} presents the requirements and architecture of the system design. Section~\ref{sec:method} details the implementation and evaluation methodology. Section~\ref{sec:eval} reports the results of the automated and human evaluation. Section~\ref{sec:discussion} discusses limitations and Section~\ref{sec:conclusion} concludes our work.
 
\section{Related Work}
\label{sec:related}
 
\subsection{Health Literacy and Readability of Patient Education Materials}
Health literacy is recognized by the World Health Organization as one of the pillars underpinning progress toward the Sustainable Development Goals, and low health literacy is treated as a global rather than a purely local concern~\cite{ayre2023shell}. A recurring finding in the literature is that patient education materials (PEMs), including those distributed directly from electronic health records, are systematically written above recommended reading levels: a study of PEMs for the five most common diagnoses at a hospital found that fewer than one in five patients read at or below the sixth-grade level typically recommended for such material, while the majority of PEMs assumed a substantially higher reading level~\cite{imoisili2017discrepancy,singh2024readability}. Automated tools such as the Sydney Health Literacy Lab (SHeLL) Health Literacy Editor address this gap by providing rule-based feedback across dimensions including readability, passive voice, lexical density, and person-centered language~\cite{ayre2023shell}. Such tools are valuable for authoring static materials but do not generate personalized responses to a patient's specific question, motivating interactive, generative alternatives.
 
\subsection{LLMs for Medical Text Simplification}
Generative AI has been explored as a means of producing more personalized and conversational health explanations. Clinician review of GPT-generated answers to diabetes self-management questions found a very high proportion to be factually acceptable~\cite{kelly2025chatbot}, suggesting that general LLMs can produce clinically reasonable content when prompted directly on well-covered topics. However, simplification remains a distinct challenge from factual accuracy. Prior work evaluating unprompted and simplification-prompted LLM output on a specific respiratory condition found reading levels remained in the 12th-to-14th grade range even after an explicit request to simplify~\cite{aydin2024scoping}, and broader reviews highlight this as a systemic limitation of general-purpose models rather than a topic-specific artefact~\cite{hembroff2025opensource}. Two general strategies have been proposed to close this gap: fine-tuning an open-source model on a custom simplification dataset~\cite{hembroff2025opensource}, and retrieval-based grounding combined with prompting~\cite{kelly2025chatbot}. This paper builds on the second strategy, adding an explicit, metric-driven validation loop that neither prior approach incorporates.
 
\subsection{Retrieval-Augmented Generation}
RAG augments an LLM's parametric knowledge with non-parametric knowledge retrieved from an external, often domain-specific, corpus~\cite{klesel2025rag}. The approach is typically implemented as a two-stage pipeline: an offline indexing stage that chunks, embeds, and stores trusted documents in a vector database~\cite{pan2024vectordb,yadav2025raprec}, and an online retrieval stage that embeds an incoming query, retrieves the most similar chunks, and augments the prompt sent to the generator~\cite{klesel2025rag}. Pure dense vector search is effective at capturing semantic intent but can under-perform on queries requiring exact keyword or negation matching~\cite{wang2025hybridsearch}; hybrid retrieval, which fuses dense similarity search with lexical methods such as BM25 (via strategies such as reciprocal rank fusion, weighted score combination, or LLM-based re-ranking), has been proposed to recover this gap~\cite{wang2025hybridsearch,riyadh2024llmvector}. The final augmentation step is itself a form of prompt engineering, in which retrieved context, the user query, and task-specific instructions (grounding constraints, fallback behavior, or output-format exemplars) are composed into a single prompt~\cite{klesel2025rag,singhal2024factcheck}.
 
\subsection{RAG in Healthcare and the Research Gap}
Within healthcare specifically, RAG has been applied to diagnostic decision support, summarization of electronic health records, and literature-grounded medical question answering~\cite{neha2025ragreview}. These applications share a common goal: reducing hallucination and grounding responses in verifiable, current sources such as clinical guidelines or repositories like PubMed~\cite{neha2025ragreview}. Despite this progress, the great majority of healthcare RAG research targets \emph{clinician}-facing accuracy rather than \emph{patient}-facing comprehensibility~\cite{ajhb2024misinfo}. This gap in the literature, factual grounding for clinicians rather than verified comprehensibility for patients, motivates the case study presented here: rather than proposing a new retrieval method, this paper examines, for a single high-burden condition, how existing RAG techniques can be paired with an explicit, auditable simplification component and evaluated against a measurable reading-level target.
 
\section{System Design}
\label{sec:design}
 
The system described in this section follows established RAG design patterns (hybrid retrieval, vector-store indexing, prompt-based augmentation); it is presented here as the concrete pipeline used for this case study, not as a novel architectural contribution. The elements of the design that are specific to this work, the diabetes-specific lay-term lexicon and the diagnostic readability validator, are described in Section~\ref{sec:method}.
 
\subsection{Requirements}

System requirements were derived from the two challenges identified in Section~\ref{sec:related}: retrieved clinical content must remain factually grounded and traceable to its source, and generated responses must reliably meet a target reading level rather than merely appearing simpler. These two concerns map onto distinct parts of the pipeline: factual grounding is addressed by the retrieval and source-attribution requirements, while readability is addressed by the generation-constraint and validation requirements. Together they define what the system must do (functional requirements) and the quality bar it must meet (non-functional requirements). For brevity within the page limit, the full functional and non-functional requirements tables that guided the design are provided as supplementary material (Tables 1 and 2 therein) rather than in the main text; in summary, the functional requirements specify that the system must ingest and index a curated corpus, retrieve relevant chunks per query, assemble a single prompt from the retrieved context and the query, constrain generation to a sixth-grade reading level, and attach the source to every response, while the non-functional requirements specify the FKGL~$\leq$~6.0 readability target, a response-time ceiling, freedom from hallucination, and secure handling of patient data.

\subsection{Architecture}

\begin{figure}[H]
\centering
\includegraphics[width=6.5cm]{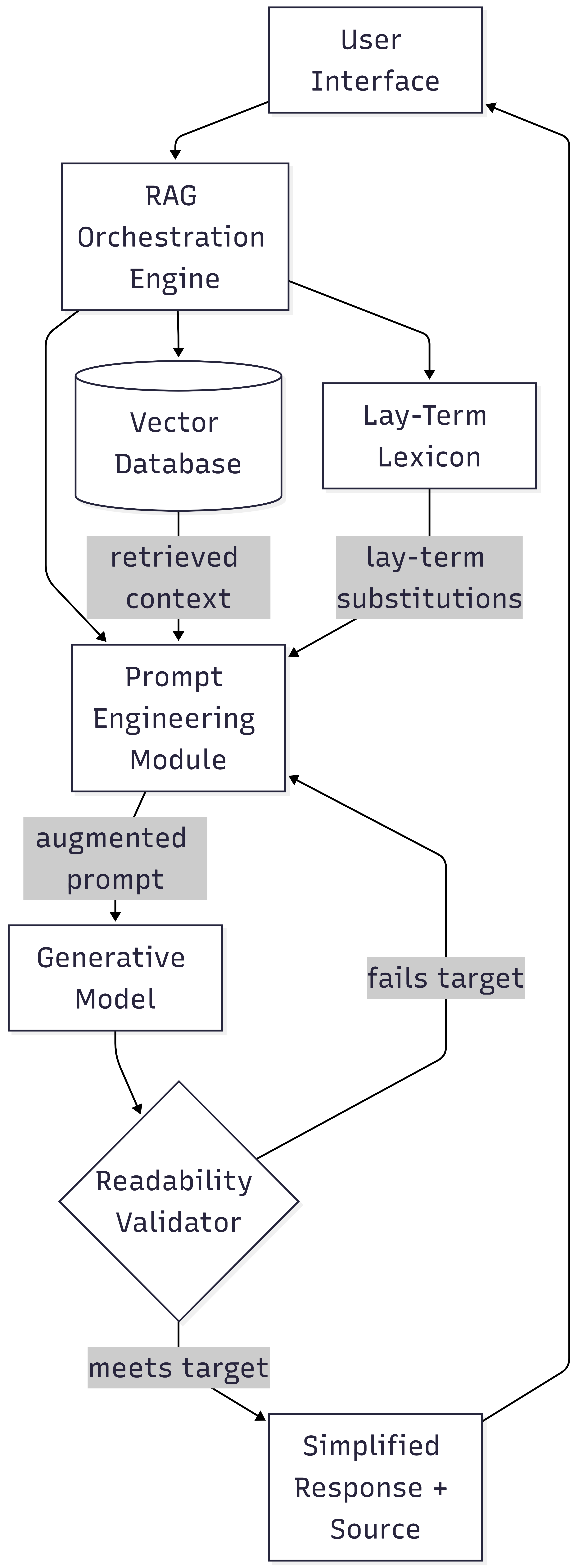}
\caption{The system architecture design of \emph{MediClear}.}
\label{fig:cap1}
\end{figure}

Figure~\ref{fig:cap1} illustrates the core components of the pipeline used in this case study. A \emph{user interface} accepts a natural-language query and displays the final simplified response together with its source. A \emph{RAG orchestration engine} coordinates retrieval, augmentation, generation, and validation. A \emph{vector database} stores embedded chunks of the trusted, dynamically retrieved knowledge base and supports similarity search. A \emph{lay-term lexicon} is a separate, static store of clinical-to-lay term substitutions (Section~\ref{sec:method}) that is not indexed or retrieved, it is always available in full. A \emph{prompt engineering module} assembles the augmented prompt sent to the generator by combining the retrieved context, the lexicon and formatting rules, and the user query. A \emph{generative model} produces the response from this augmented prompt. Finally, a \emph{readability validator} computes FKGL and SMOG scores for each candidate response; if the target is not met, the orchestration engine, via the prompt engineering module, constructs a targeted re-prompt rather than a generic ``simplify further'' instruction.

\subsection{Workflow}
Consider a patient who has just been diagnosed with diabetes and asks, ``What is diabetes?''. The query is embedded and matched against the vector database, which returns the top four semantically relevant chunks together with their source (e.g., a national diabetes association). The orchestration engine assembles a structured prompt consisting of the system instructions, the retrieved chunks, and the original query, and sends it to the generator. The resulting response is scored by the readability validator. If both FKGL and SMOG fall at or below the target of 6.0, the response is returned to the user with its source attribution. If not, the engine issues a targeted re-prompt (see Appendix C of the supplementary material for the full prompt templates), for example instructing the model to shorten sentences or replace specific multi-syllable terms, and the cycle repeats up to a bounded number of times before the best available response is returned.
 
\section{Methodology}
\label{sec:method}
 
\subsection{Implementation}
The system, named MediClear, was implemented using the LangChain framework to orchestrate the indexing and retrieval-generation pipelines. OpenAI's GPT-4o-mini was used as the generative model for its strong instruction-following behavior, and the \texttt{text-embedding-3-small} model was used to embed both the indexed corpus and incoming queries. Facebook AI Similarity Search (FAISS) was used as the vector store. Retrieval was implemented as a hybrid search: a lexical component using the BM25 algorithm (via the \texttt{rank\_bm25} library) retrieves the top four chunks by keyword match, while a dense component retrieves the top four chunks by embedding similarity; the two result sets are merged before being passed to the generator. Readability was computed using the \texttt{pyphen} syllable-counting library, adapted with a custom bullet-to-sentence preprocessing step to correct a known failure mode in which list-formatted responses were mis-scored by standard readability libraries. The user interface was built with Streamlit.
 
\subsection{Data Source Curation}
Diabetes was selected as the evaluation domain given its prevalence and the degree of self-management knowledge it demands from patients~\cite{agboluaje2025}; management guidance in this domain is also mature and well-documented, which supports construction of a reliable knowledge base. The non-parametric memory of the system was built from a curated set of trusted, patient-facing and clinical-guidance sources, including a national diabetes association, the World Health Organization, the American Diabetes Association Standards of Care, national institute patient-education materials, and a national institute of health and welfare. These sources were selected for their evidentiary standing and their established use as reference material for diabetes patients and clinicians alike.
 
Indexing followed a standard offline pipeline: documents were ingested via a document loader and normalised into a standard format; text was split into 1000-character chunks with a 200-character overlap using a recursive character splitter, chosen to preserve semantic continuity across chunk boundaries; each chunk was embedded using the \texttt{text-embedding-3-small} model; and the resulting vectors were stored in the FAISS index.
 
\subsection{Simplification Mechanism}
A key design decision separates the roles of retrieval and simplification, and this section makes that separation, and its limitations, explicit.
 
\textbf{Retrieval component (dynamic, per-query).} The hybrid retriever is responsible only for surfacing the most relevant raw chunks from the trusted, indexed corpus (Section~4.2); these chunks are frequently written at a clinical reading level and are retrieved fresh for every query via the FAISS vector store and BM25 index described above.
 
\textbf{Lay-term lexicon (static, not retrieved).} Simplification vocabulary is handled separately from retrieval. It takes the form of a fixed, manually authored lexicon of roughly 30 clinical-to-lay term substitutions (e.g., ``glycated haemoglobin'' to ``blood sugar test'', ``hyperglycaemia'' to ``high blood sugar'', ``myocardial infarction'' to ``heart attack''; the full list is given in Appendix B of the underlying project report). This directly answers three design questions the lexicon raises. \emph{Source:} it was compiled manually by the author, by reviewing the vocabulary that recurred across the retrieved diabetes corpus (Diabetes Australia, WHO, ADA Standards of Care, NIDDK materials, AIHW) and identifying the multi-syllable clinical terms that most frequently caused non-simplified responses to exceed the target reading level; it was not mined automatically or learned from data. \emph{Usage:} it is supplied entirely as static content inside the constrained system prompt given to the generator for the simplified-response branch and is \emph{not} embedded, indexed, or retrieved through the vector database, which holds only the dynamic clinical-content corpus (Section~4.2). \emph{Linking mechanism:} the mapping is a deterministic, one-to-one substitution table, not a learned or embedding-based association; the LLM is instructed to apply these substitutions, together with general word- and sentence-length rules, when rephrasing retrieved content. This was chosen for verifiability, every substitution can be checked against the prompt, at the cost of not scaling automatically to other conditions, a trade-off discussed further in Section~\ref{sec:discussion}.
 
In addition to the lexicon, the constrained prompt also specifies a maximum sentence length and formatting rules for different question types (short paragraphs for definitional questions, bulleted lists for symptom questions). Together, retrieval and simplification are optimized independently: the retrieved context anchors the response in verified source material, while the lexicon-driven prompt governs \emph{how} that material is reworded.
 
If a generated response still exceeds the FKGL~6.0 threshold after this prompt is applied, a diagnostic retry loop is triggered. Rather than issuing a generic re-prompt, the system computes the FKGL score, determines whether the dominant cause is excessive sentence length or excessive word complexity (i.e., use of a term outside the lexicon), and constructs a targeted correction instruction accordingly. This process is repeated for up to three attempts before the best available response is returned.
 
\subsection{Evaluation Design}
Two evaluation strategies were used: an automated readability evaluation and a human evaluation.
 
\textbf{Automated evaluation.} Three question sets were constructed to probe performance across a range of clinical complexity: a set of 30 general diabetes questions spanning definitions, management, and complications, representative of a newly diagnosed patient's likely queries; a subset of 10 \emph{easy} questions covering basic patient-level knowledge (definitions, symptoms, lifestyle advice); and a subset of 10 \emph{hard} questions covering clinically complex topics such as disease mechanisms, complications, and medication effects. Questions were classified as easy or hard using three criteria: conceptual complexity, terminology density of the retrieved source context, and the degree of prior clinical knowledge assumed. A question was classified as high complexity if it satisfied at least two of the three criteria. For every question, two responses were generated for comparison: a \emph{non-simplified baseline}, in which retrieved content was passed to the generator with only a grounding instruction, and a \emph{simplified} response, in which the same retrieved content was passed with the full lexicon-based constrained prompt described above. Both response types were scored using the FKGL,
\begin{equation}
\text{FKGL} = 0.39 \left(\frac{\text{words}}{\text{sentences}}\right) + 11.8\left(\frac{\text{syllables}}{\text{words}}\right) - 15.59,
\end{equation}
and SMOG index,
\begin{equation}
\text{SMOG} = 3 + \sqrt{\text{polysyllables} \times \frac{30}{\text{sentences}}},
\end{equation}
both standard, formula-based readability measures in which lower scores indicate easier text~\cite{readable2025flesch,readable2025smog}.
 
\textbf{Human evaluation.} An in-application feedback form was presented to participants after each simplified response. Participants first indicated whether they had a medical or healthcare background, then rated three dimensions on a five-point scale: ease of understanding, trustworthiness, and overall satisfaction, and finally indicated whether they would use the tool again (Yes / Maybe / No). Responses were logged automatically for analysis.
 
\section{Evaluation}
\label{sec:eval}
 
\subsection{Automated Readability Evaluation}
Table~\ref{tab:general} summarizes results across the 30 general diabetes questions. Non-simplified responses averaged an FKGL of 7.08 (SD 2.49, range 2.78 to 12.25), confirming that ungrounded, unconstrained generation produces reading levels well above the target even for common patient questions. Simplified responses averaged an FKGL of 4.27 (SD 1.02, range 1.40 to 6.19), a 39.8\% reduction, with 29 of 30 questions (96.7\%) meeting the FKGL~$\leq$~6.0 target. The single question that did not meet the target (FKGL 6.19) required the response to cover multiple sub-topics simultaneously. Prior development-stage testing indicated the diagnostic retry loop is triggered for roughly half of queries on average (approximately 0.8 retries per question), indicating that the initial simplified prompt alone is insufficient to reliably hit the target for every question and that the diagnostic retry mechanism is a necessary, not merely incidental, part of the design.
 
\begin{table}[t]
\centering
\caption{Readability results across the 30 general diabetes questions (mean $\pm$ SD).}
\label{tab:general}
\begin{tabular}{@{}lc@{}}
\toprule
\textbf{Metric} & \textbf{Value} \\
\midrule
Total questions evaluated & 30 \\
Non-simplified FKGL (mean $\pm$ SD) & 7.08 $\pm$ 2.49 \\
Simplified FKGL (mean $\pm$ SD) & 4.27 $\pm$ 1.02 \\
Mean FKGL improvement & 39.8\% \\
Questions meeting FKGL $\leq$ 6.0 & 29/30 (96.7\%) \\
\bottomrule
\end{tabular}
\end{table}
 
\subsection{Comparative Evaluation: Easy versus Hard Questions}
Table~\ref{tab:comparison} compares performance on the easy and hard question subsets. For easy questions, non-simplified responses averaged an FKGL of 7.77 (SD 2.17), falling to 4.04 (SD 0.96) after simplification, a 48.0\% improvement, with all 10 questions (100\%) meeting the target and no retries required. For hard questions, non-simplified responses averaged a substantially higher FKGL of 11.79 (SD 2.76), confirming that clinically complex retrieved content is inherently harder to read; yet simplified responses still achieved a mean FKGL of 5.01 (SD 0.66), a larger relative improvement of 57.5\%, with all 10 questions (100\%) meeting the target and a low mean retry count of 0.2 (2 of 10 questions required one retry).
 
\begin{table}[t]
\centering
\caption{Comparison of readability metrics between easy and hard question sets (mean $\pm$ SD).}
\label{tab:comparison}
\begin{tabular}{@{}lcc@{}}
\toprule
\textbf{Metric} & \textbf{Easy questions} & \textbf{Hard questions} \\
\midrule
Non-simplified FKGL & 7.77 $\pm$ 2.17 & 11.79 $\pm$ 2.76 \\
Simplified FKGL & 4.04 $\pm$ 0.96 & 5.01 $\pm$ 0.66 \\
Mean FKGL improvement & 48.0\% & 57.5\% \\
Target met (FKGL $\leq$ 6.0) & 10/10 (100\%) & 10/10 (100\%) \\
Mean retries per question & 0.0 & 0.2 \\
\bottomrule
\end{tabular}
\end{table}
 
Two findings follow from this comparison. First, the combined target-achievement rate across both sets is 100\% (20/20), indicating that the simplification pipeline is robust to variation in source complexity rather than being tuned narrowly to easy cases. Second, the \emph{relative} improvement is larger for hard questions (57.5\%) than for easy questions (48.0\%), meaning the mechanism provides proportionally greater benefit exactly where source material is most technical, the scenario in which health-literacy support is most needed. The corresponding rise in retries for hard questions (0.0 to 0.2) confirms that the diagnostic retry loop, while rarely needed for either subset, is exercised somewhat more for technically dense content. The consistently lower standard deviation of simplified FKGL relative to non-simplified FKGL in both subsets (e.g., hard: SD 0.66 vs.\ 2.76) further indicates that the constrained prompt and lexicon narrow the spread of reading levels, not just the mean, making output difficulty more predictable regardless of question complexity.
 
\subsection{Human Evaluation}
A total of 49 rated responses were collected from 10 participants (5 with, and 5 without, a self-reported medical or healthcare background). Table~\ref{tab:human} summarizes the results. Ease of understanding received the highest mean rating (4.76/5), with 95.9\% of ratings at 4 or 5, consistent with the automated finding that FKGL-targeted responses are perceived as substantially easier to read. Trustworthiness received a mean rating of 4.02/5 (77.6\% rated 4 or 5); the somewhat lower and more dispersed trustworthiness scores may partly reflect participant awareness that responses were AI-generated rather than clinician-reviewed, a concern the source-attribution mechanism is designed to mitigate. Overall satisfaction averaged 4.33/5 (77.6\% rated 4 or 5); the modest gap between ease-of-understanding and satisfaction scores is consistent with a known trade-off in text simplification, whereby increased readability can come at the cost of some loss of clinical nuance~\cite{aydin2024scoping}, and with occasional fallback responses returned when the knowledge base lacked sufficient context for a query.
 
\begin{table}[t]
\centering
\caption{Human evaluation ratings across all participants ($n=49$ responses).}
\label{tab:human}
\begin{tabular}{@{}lccc@{}}
\toprule
\textbf{Rating Dimension} & \textbf{Mean} & \textbf{Mode} & \textbf{Score 4-5 (\%)} \\
\midrule
Ease of Understanding & 4.76 & 5 & 95.9\% \\
Trustworthiness & 4.02 & 4 & 77.6\% \\
Overall Satisfaction & 4.33 & 5 & 77.6\% \\
\bottomrule
\end{tabular}
\end{table}
 
Disaggregating by background, non-medical participants rated ease of understanding notably higher (4.78) than participants with a medical background (3.50), an expected outcome since simplified responses are designed for a lay audience and may read as under-informative to clinically trained readers. Trustworthiness ratings were, by contrast, nearly identical across groups (4.00 for medical, 4.02 for non-medical participants), suggesting that the retrieval-grounding and source-attribution mechanism produces a consistent perception of reliability independent of the reader's background. Finally, 45 of 49 responses (91.8\%) indicated the participant would use the tool again, 4 (8.2\%) indicated ``Maybe'', and none indicated ``No'', suggesting strong overall acceptance of the simplified-language-plus-source-attribution approach.
 
\subsection{Summary Across Evaluation Conditions}
Table~\ref{tab:summary} consolidates the automated results across all three evaluation conditions. Across every condition, the system reduces mean FKGL to within the sixth-grade target, and the target-achievement rate is uniformly high (96.7-100\%) regardless of question difficulty. Combined with the human-evaluation results, this indicates that grounding retrieval in a trusted corpus and separately optimizing for readability through a diagnostic, closed-loop mechanism is an effective strategy for improving both the measured and perceived accessibility of AI-generated health information.
 
\begin{table}[t]
\centering
\caption{Overall quantitative results across all evaluation conditions (mean $\pm$ SD).}
\label{tab:summary}
\begin{tabular}{@{}lccc@{}}
\toprule
\textbf{Metric} & \textbf{30 General} & \textbf{Easy (10)} & \textbf{Hard (10)} \\
\midrule
Non-simplified FKGL & 7.08 $\pm$ 2.49 & 7.77 $\pm$ 2.17 & 11.79 $\pm$ 2.76 \\
Simplified FKGL & 4.27 $\pm$ 1.02 & 4.04 $\pm$ 0.96 & 5.01 $\pm$ 0.66 \\
Mean improvement & 39.8\% & 48.0\% & 57.5\% \\
Target met (FKGL $\leq$ 6.0) & 96.7\% & 100\% & 100\% \\
\bottomrule
\end{tabular}
\end{table}
 
\section{Discussion and Limitations}
\label{sec:discussion}
 
As a case study, the results speak most directly to what is required in practice, not merely in principle, to close the gap identified in prior literature, where a single generic simplification request was shown to leave LLM output well above target reading levels~\cite{aydin2024scoping}. Two design choices appear central to the improvement observed here: separating a manually curated, domain-specific lexicon from the retrieval step, so that vocabulary control does not depend on what happens to be retrieved; and enforcing the reading-level target through a machine-verifiable, diagnostic retry mechanism rather than a single unverified simplification instruction. The larger relative improvement observed for clinically complex questions (57.5\% for hard vs.\ 48.0\% for easy) further suggests that this mechanism scales with source difficulty rather than being limited to easy cases, which is an important property for a tool intended to serve patients across the full range of clinical topics they may encounter. The reduced standard deviation of simplified FKGL relative to non-simplified FKGL across all three conditions also indicates the approach narrows variability, not only the average reading level, making output difficulty more consistent and predictable for the reader.
 
Several limitations should be acknowledged, most of which follow from the case-study, single-condition scope of this work rather than from the general RAG pipeline. First, and most significantly, the lay-term lexicon is manually curated and diabetes-specific: it was built by hand for roughly 30 terms recurring in this corpus, and nothing about its construction generalizes automatically to another condition. Extending the approach would require either repeating this manual process per condition or replacing it with an automated mechanism, for example embedding-based similarity between clinical and lay terms drawn from an existing plain-language medical glossary, which is left as a concrete direction for future work. Second, the knowledge base is currently restricted to a small number of trusted diabetes-focused sources, constraining topic coverage. Third, the human evaluation involved only 10 participants (49 rated responses); a larger and more demographically varied sample is needed for robust generalisable conclusions. Fourth, the SMOG index requires a minimum sentence count to return a valid score, and several concise simplified responses fell below this threshold, so FKGL served as the primary metric. Finally, because the evaluation is confined to a single clinical domain with a hand-built lexicon, the improvements reported here should be read as evidence of what this specific, transparent approach can achieve for one condition, not as a general claim about RAG-based health-literacy systems.
 
\section{Conclusion}
\label{sec:conclusion}
 
This paper presented \emph{MediClear}, a case study in using established RAG techniques, a manually curated lay-term lexicon, and a diagnostic, closed-loop readability validator to make diabetes-related clinical information accessible to patients. The RAG pipeline itself follows standard practice; the paper's contribution is the transparent documentation and evaluation of how factual grounding and readability can be controlled as separate, auditable components, and what this achieves in practice for one high-burden chronic condition. Applying this approach, the system reduced the mean FKGL of generated diabetes-related responses from a non-simplified baseline of 7.08-11.79 (SD 2.17-2.76) to a simplified 4.04-5.01 (SD 0.66-1.02), meeting a sixth-grade readability target for 96.7-100\% of evaluated questions depending on clinical complexity. A human evaluation corroborated these automated findings, with participants rating simplified responses as substantially easier to understand (mean 4.76/5) and expressing a strong willingness to reuse the tool (91.8\%), while trustworthiness ratings remained consistent regardless of participants' medical background. 

\subsection{Future Work}
The lexicon underlying these results was hand-built for diabetes, so the central open question this case study raises is how such a lexicon could be constructed automatically, for example, via embedding-based similarity between clinical and lay terminology, so that this approach generalizes to other conditions without repeating the manual process. Future work will pursue this direction, extend the knowledge base and evaluation to additional chronic conditions, incorporate a larger and more diverse participant pool, and investigate fine-tuning as a complementary strategy to prompt-based simplification.

\subsubsection*{AI Usage Disclosure.} In accordance with the conference's transparency requirements, the authors disclose that generative AI tools were used during the preparation of this work. A general-purpose AI assistant was used to help structure literature review headings and to assist with debugging application code. A research-assistant tool integrated into an institutional library search platform was used to help identify a small number of candidate reference papers. An AI coding/writing assistant was used to help draft evaluation questions and system prompts used within the described application, and to assist in diagnosing technical errors during development. A separate general-purpose AI assistant was used to assist in correcting diagram source code. Grammar-checking software was used to assist with proofreading. All AI-assisted output was reviewed, verified, and edited by the authors, who take full responsibility for the accuracy, integrity, and originality of the content presented in this paper.
 
\bibliographystyle{ieeetr}
\bibliography{references}
 
\end{document}


\title{Appendix: Medical Knowledge Simplification for Patients in the Era of LLMs: A Case Study on Diabetes}

\author{Anonymous Author(s)}
\institute{Anonymous Institution(s)}

\maketitle

\section*{Overview}
This supplementary material accompanies the main submission. It contains: (A) the full functional and non-functional requirements tables referenced from the System Design section of the main paper; (B) a sequence diagram detailing the message-level interaction flow for a single patient query; (C) the constrained system prompts used to generate the non-simplified and simplified responses, together with the retry-loop prompt template; and (D) the full evaluation question sets (general, easy, and hard) used for the automated readability evaluation.

\section*{A. Requirements}
Table~\ref{tab:fr} and Table~\ref{tab:nfr} list the functional and non-functional requirements that guided the design of the case-study pipeline described in the main paper.

\begin{table}[h]
\centering
\caption{Functional requirements (FR).}
\label{tab:fr}
\begin{tabular}{@{}p{0.06\textwidth}p{0.78\textwidth}p{0.1\textwidth}@{}}
\toprule
\textbf{ID} & \textbf{Description} & \textbf{Priority} \\
\midrule
FR1 & Ingest a curated corpus of evidence-based medical documents into a knowledge base. & High \\
FR2 & Preserve source and page metadata for each indexed chunk. & High \\
FR3 & Store chunk embeddings in a vector database. & High \\
FR4 & Accept natural-language patient queries and convert them to query vectors. & High \\
FR5 & Retrieve the top-$k$ most relevant chunks via similarity search. & High \\
FR6 & Assemble a single prompt from the system instructions, retrieved context, and user query. & High \\
FR7 & Constrain generation strictly to retrieved content. & High \\
FR8 & Constrain generation to a sixth-grade target reading level. & High \\
FR9 & Attach the originating source to every generated response. & Medium \\
\bottomrule
\end{tabular}
\end{table}

\begin{table}[h]
\centering
\caption{Non-functional requirements (NFR).}
\label{tab:nfr}
\begin{tabular}{@{}p{0.08\textwidth}p{0.76\textwidth}p{0.1\textwidth}@{}}
\toprule
\textbf{ID} & \textbf{Description} & \textbf{Priority} \\
\midrule
NFR1 & Generated text must achieve an FKGL score of 6.0 or below. & High \\
NFR2 & End-to-end response latency should not exceed 10 seconds. & Low \\
NFR3 & Generated text must remain factual and free of hallucination. & High \\
NFR4 & The system must handle patient data securely. & High \\
\bottomrule
\end{tabular}
\end{table}

\section*{B. Sequence Diagram}

\begin{figure}[H]
    \centering
    \rotatebox{90}{%
        \includegraphics[height=0.70\linewidth, keepaspectratio]{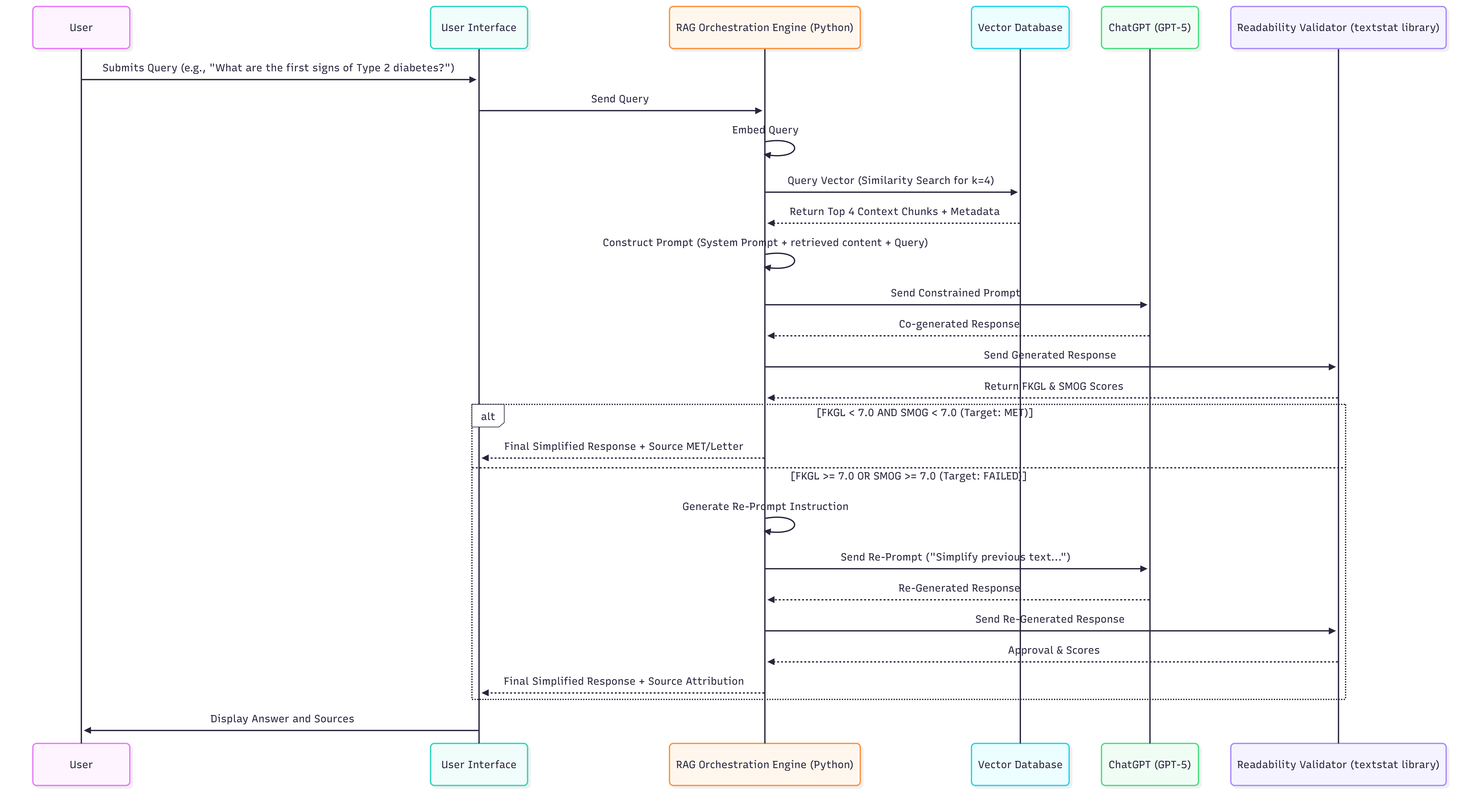}%
    }%
    \caption{Sequence diagram for a single patient query, from initial submission through retrieval, augmented generation, readability validation, the conditional re-prompt loop, and final display.}
    \label{fig:seq}
\end{figure}


Figure~\ref{fig:seq} details the interaction sequence corresponding to the workflow described in the main paper, showing message passing between the user, the orchestration engine, the vector database, the generative model, and the readability validator for a single query, including the closed-loop re-prompt path.

\section*{C. System Prompts}
\label{app:prompts}
This section reproduces the constrained system prompts used to generate the non-simplified and simplified responses described in Section 4 of the main paper, together with the retry-loop prompt template used by the diagnostic readability validator.

\subsection*{C.1 Non-Simplified Prompt}
Used to generate the baseline response for comparison; it constrains the generator to the retrieved context but applies no readability constraint.

\begin{verbatim}
NON_SIMPLIFIED_PROMPT = """You are an AI Medical Information Retriever.
Your task is to act as a conversational interface for a professional
user seeking specific medical information.

You will be given a QUERY and CONTEXT chunks from trusted medical
documents.

Rules:
1. Answer using ONLY the CONTEXT provided. No outside knowledge.
2. If the answer is not in CONTEXT say:
   "I'm sorry, I could not find that in the provided medical
   documents."
3. Tone must be objective, formal and academic.
4. End with: Source: [name of actual source document used]"""
\end{verbatim}

\subsection*{C.2 Simplified Prompt}
Used to generate the lexicon-constrained, sixth-grade-target response described in Section 4.3 of the main paper. The word-substitution rules shown here are a representative excerpt of the approximately 30-entry lexicon; the sentence-, format-, and example-rules are reproduced in full.

\begin{verbatim}
SIMPLIFIED_PROMPT = """You are a health educator writing patient
information.

ONLY use information from the CONTEXT provided. Do not add outside
knowledge.

YOUR TARGET: Flesch-Kincaid Grade Level of 6 or below.

The FKGL formula is:
FKGL = 0.39 x (words per sentence) + 11.8 x (syllables per word)
       - 15.59

To score 6 or below you must:
- Keep average words per sentence below 14
- Keep average syllables per word below 1.5
- This means AT LEAST 70% of your words must be 1 syllable

WORD RULES - most important factor. Choose the shortest word
every time:
- "blood sugar" not "glucose"
- "heart" not "cardiovascular"
- "signs" not "symptoms"
- "check" not "monitor"
- "drugs" or "medicine" not "medication"
- "care" not "treatment"
- "high blood sugar" not "hyperglycemia"
- "low blood sugar" not "hypoglycemia"
- "harm" not "complication"
- "shots" not "injections"
- "tired" not "fatigue"
- "linked" not "associated"
- "about" not "approximately"
- "need" not "require"
- "get" not "develop"
- "give" not "provide"
- "make" not "produce"
- "stop" not "prevent"
- "keep" not "maintain"
- "rise" not "increase"
- "fall" not "decrease"
- "use" not "utilise"
- "help" not "assist"
- "but" not "however"
- "so" not "therefore"
- "also" not "additionally"
- "food" not "nutrition"
- "long-term" not "chronic"
- "people" not "individuals"
- "often" not "frequently"
- "illness" or "condition" not "disease"
- "body" not "organism"

SENTENCE RULES:
- Every sentence must be under 14 words. Count every word before
  you write it.
- Never connect two ideas with "which", "although", "however" in
  one sentence.
- Split long ideas into two short sentences.

FORMAT RULES:
- For "What is" questions: write 2 to 3 short paragraphs
- For "What are the signs/symptoms" questions: use bullet points
- For "How is it managed" questions: use a mix of both
- Start with: "The documents state that..."
- End with: Source: [actual source document name]

GOOD EXAMPLE for "What is type 2 diabetes?":
The documents state that type 2 diabetes is a long-term illness.
It affects how the body uses blood sugar. The body does not use
insulin well. Blood sugar gets too high. This can harm the heart,
eyes and kidneys over time.

There are ways to keep it in check. Good food and daily walks
help. Some people also need drugs to control their blood sugar.

Source: Diabetes Australia

GOOD EXAMPLE for "What are the signs of diabetes?":
The documents state that diabetes has a number of common signs.

- Feeling very tired each day
- Being very thirsty often
- Needing to urinate more than normal
- Blurred sight
- Slow healing cuts
- Losing weight without trying

Source: NHS UK

If the answer is not in the CONTEXT say:
"I could not find this in the medical documents provided." """
\end{verbatim}

\subsection*{C.3 Retry-Loop Prompt Template}
Constructed dynamically by the orchestration engine when a candidate response exceeds the FKGL~6.0 threshold; \texttt{\{fkgl\}} and \texttt{\{issues\}} are populated at run time with the measured score and the specific diagnostic cause (excessive sentence length, excessive word complexity, or both).

\begin{verbatim}
RETRY_PROMPT_TEMPLATE = """The previous response scored FKGL
{fkgl:.1f} - above the target of 6.0.

The specific problems are:
{issues}

Rewrite the ENTIRE response from scratch. Apply these fixes:
- Break any sentence over 10 words into two separate sentences
- Replace every multi-syllable word using the word list in the
  system prompt
- Keep every bullet point under 10 words
- Do NOT add any new information - only use what was in the
  original context"""
\end{verbatim}

\section*{D. Evaluation Question Sets}
\label{app:questions}
This section lists the three question sets used for the automated readability evaluation described in Section 4.4 and 5 of the main paper.

\subsection*{D.1 General Question Set (n = 30)}
\textit{Note:} 29 of the 30 questions in this set are reproduced below from the project's source material; the wording of the remaining question was not legible in the archived source document and is omitted here rather than reconstructed. The corresponding data point is nonetheless included in the raw per-question results used to compute the statistics reported in the main paper.

\begin{verbatim}
Basic knowledge
1. What is diabetes?
2. What is prediabetes?
3. Are all types of diabetes the same?
4. What happens when people with diabetes do not receive insulin?
5. What causes type 1 diabetes?
6. What causes type 2 diabetes?
7. What is gestational diabetes?
8. What is insulin and what does it do?
9. How common is diabetes in Australia?
10. What is the difference between type 1 and type 2 diabetes?

Symptoms
11. What are the symptoms of type 2 diabetes?
12. What are the signs of prediabetes?
13. What are the signs of low blood sugar?
14. What are the signs of high blood sugar?
15. What are the early warning signs of diabetes?

Management
16. How is type 2 diabetes managed?
17. How is type 1 diabetes managed?
18. How can I check my blood sugar levels?
19. What foods should people with diabetes eat?
20. How does exercise help people with diabetes?
21. What medicines are used to treat type 2 diabetes?
22. What is an HbA1c test?
23. How often should people with diabetes see a doctor?

Complications
24. What are the long-term complications of diabetes?
25. How does diabetes affect the kidneys?
26. How does diabetes affect the eyes?
27. How does diabetes affect the heart?
28. What is diabetic ketoacidosis?

Prevention and risk
29. Can type 2 diabetes be prevented?
30. Who is at risk of developing type 2 diabetes?
\end{verbatim}

\subsection*{D.2 Easy Question Set (n = 10)}
Basic, patient-level knowledge questions used for the comparative evaluation in Section 5.2 of the main paper.

\begin{verbatim}
1. What is diabetes?
2. What is prediabetes?
3. What causes type 2 diabetes?
4. What are the symptoms of type 2 diabetes?
5. What is insulin and what does it do?
6. How is type 2 diabetes managed?
7. What foods should people with diabetes eat?
8. Can type 2 diabetes be prevented?
9. What are the early warning signs of diabetes?
10. How does exercise help people with diabetes?
\end{verbatim}

\subsection*{D.3 Hard Question Set (n = 10)}
Clinically complex questions used for the comparative evaluation in Section 5.2 of the main paper.

\begin{verbatim}
1. What are the long-term complications of uncontrolled diabetes
   on the cardiovascular system?
2. How does diabetes affect kidney function and what are the
   warning signs of diabetic nephropathy?
3. What is the relationship between obesity and the development
   of type 2 diabetes?
4. How does gestational diabetes affect both the mother and the
   baby during pregnancy?
5. What are the differences in causes, symptoms and management
   between type 1 and type 2 diabetes?
6. How does hypoglycemia occur in people with diabetes and what
   are the recommended treatments?
7. What role does the HbA1c test play in monitoring long-term
   blood sugar control in diabetes?
8. How does diabetes contribute to nerve damage and what are the
   symptoms of diabetic neuropathy?
9. What are the recommended blood glucose targets for people with
   type 2 diabetes and why?
10. How does diabetes affect wound healing and why are foot
    complications particularly serious?
\end{verbatim}